\documentclass[conference]{IEEEtran}
\IEEEoverridecommandlockouts
\usepackage{cite}
\usepackage{amsmath,amssymb,amsfonts}
\usepackage{graphicx}
\usepackage{textcomp}
\usepackage[dvipsnames]{xcolor}

\definecolor{mygreen}{RGB}{0, 128, 0}
\definecolor{myblue}{RGB}{51, 51, 255}
\definecolor{myred}{RGB}{255, 67, 20}

\usepackage{booktabs}
\usepackage{pifont}
\usepackage{algorithm}
\usepackage{algpseudocode}

\newcommand{\cmark}{\ding{51}}%
\newcommand{\xmark}{\ding{55}}%

\usepackage{amsmath,amsfonts,bm}

\def\eqref#1{equation~\ref{#1}}

\def\1{\bm{1}}

\def\ve{{\bm{e}}}

\def\vh{{\bm{h}}}

\def\vm{{\bm{m}}}

\def\vo{{\bm{o}}}

\def\vq{{\bm{q}}}

\def\vs{{\bm{s}}}
\def\vt{{\bm{t}}}
\def\vu{{\bm{u}}}

\def\vx{{\bm{x}}}
\def\vy{{\bm{y}}}

\def\mB{{\bm{B}}}

\def\mW{{\bm{W}}}

\DeclareMathAlphabet{\mathsfit}{\encodingdefault}{\sfdefault}{m}{sl}
\SetMathAlphabet{\mathsfit}{bold}{\encodingdefault}{\sfdefault}{bx}{n}

\def\sD{{\mathbb{D}}}

\def\sW{{\mathbb{W}}}

\def\BibTeX{{\rm B\kern-.05em{\sc i\kern-.025em b}\kern-.08em
    T\kern-.1667em\lower.7ex\hbox{E}\kern-.125emX}}
\begin{document}

\title{TESS: A Scalable \underline{Te}mporally and \underline{S}patially Local Learning Rule for \underline{S}piking Neural Networks
\thanks{This work was supported in part by the Center for Co-design of Cognitive Systems (CoCoSys), one of the seven centers in JUMP 2.0, a Semiconductor Research Corporation (SRC) program, in part by the Department of Energy (DoE), and in part by the NSF AccelNet NeuroPAC Fellowship Program.}
}
\author{\IEEEauthorblockN{Marco~P.~E.~Apolinario\textsuperscript{1,2}, Kaushik~Roy\textsuperscript{1}, Charlotte~Frenkel\textsuperscript{2}}
\IEEEauthorblockA{\textsuperscript{1}\textit{School of Electrical and Computer Engineering, Purdue University, Indiana, USA} \\
\textsuperscript{2}\textit{Microelectronics Department, Delft University of Technology, Delft, NL}\\
mapolina@purdue.edu, kaushik@purdue.edu, c.frenkel@tudelft.nl}
}

\maketitle

\begin{abstract}
The demand for low-power inference and training of deep neural networks (DNNs) on edge devices has intensified the need for algorithms that are both scalable and energy-efficient. 
While spiking neural networks (SNNs) allow for efficient inference by processing complex spatio-temporal dynamics in an event-driven fashion, training them on resource-constrained devices remains challenging due to the high computational and memory demands of conventional error backpropagation (BP)-based approaches. 
In this work, we draw inspiration from biological mechanisms such as eligibility traces, spike-timing-dependent plasticity, and neural activity synchronization to introduce TESS, a \underline{te}mporally and \underline{s}patially local learning rule for training \underline{S}NNs. 
Our approach addresses both temporal and spatial credit assignments by relying solely on locally available signals within each neuron, thereby allowing computational and memory overheads to scale linearly with the number of neurons, independently of the number of time steps. 
Despite relying on local mechanisms, we demonstrate performance comparable to the backpropagation through time (BPTT) algorithm, within $\sim1.4$ accuracy points on challenging computer vision scenarios relevant at the edge, such as the IBM DVS Gesture dataset, CIFAR10-DVS, and temporal versions of CIFAR10, and CIFAR100. 
Being able to produce comparable performance to BPTT while keeping low time and memory complexity, TESS enables efficient and scalable on-device learning at the edge.
\end{abstract}


\section{Introduction}\label{sec:introduction}
With the increasing ubiquity of low-power electronic devices and the rapid advances in artificial intelligence, particularly in deep neural networks (DNNs), there has been a growing interest in bringing intelligence to the edge \cite{Abadade2023ATinyML, Singh2023EdgeSurvey, Tsoukas2024ATinyML}. 
The conventional approach, known as offline training, involves training DNNs in the cloud, where computational resources are abundant, and subsequently deploying the trained models on edge devices. 
However, certain use cases, such as those involving privacy concerns or the need for real-time model adaptation, render the offline approach unsuitable.
In these scenarios, an on-device learning paradigm is essential.
This approach requires the development of energy-efficient models and DNN learning rules that operate within the constraints of edge devices \cite{Frenkel2022ReckOn:Timescales, Lin2022On-DeviceMemory}. 

\begin{figure}[!t]
\centering
\includegraphics[width=\columnwidth]{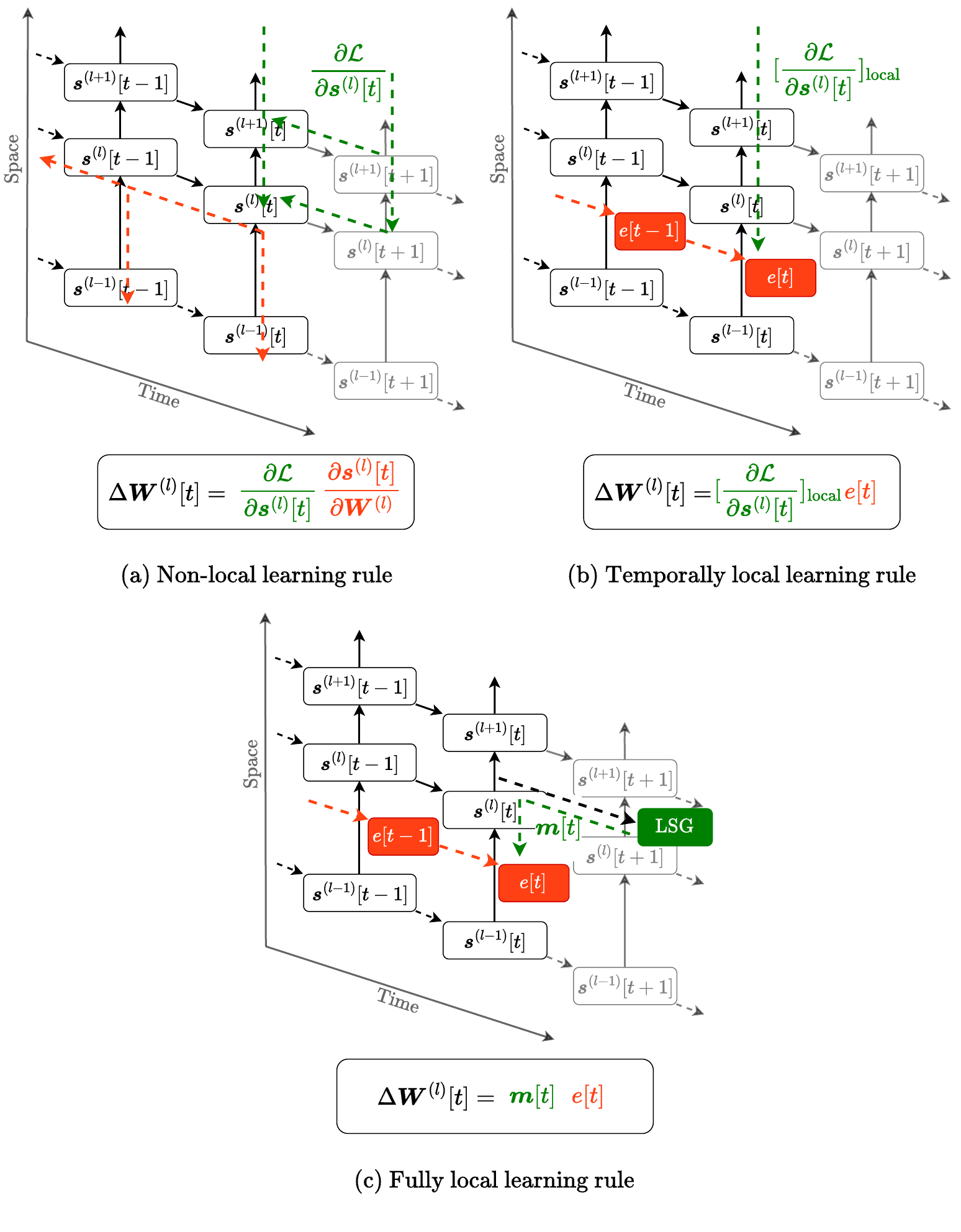}
\caption{Comparison of three learning rule strategies for training a recurrent model with state variable $\vs[t]$.  
(a) Non-local learning method (e.g., BPTT): Both spatial and temporal credit assignment problems are solved by propagating errors through time and space (layers).  
(b) Temporal local method: Temporal credit assignment is addressed using {\color{myred}eligibility traces ($\ve[t]$)}, which are auxiliary variables that track the history of neural activity. These traces are modulated by a {\color{mygreen}learning signal ($[\frac{\partial\mathcal{L}}{\partial\vs^{(l)}[t]}]_{\text{local}}$)}, which propagates errors across layers but not through time.  
(c) Fully local method (e.g., our proposed method TESS): In addition to eligibility traces, the {\color{mygreen}learning signal ($\vm[t]$)} is generated locally, addressing both spatial and temporal credit assignment entirely within the local context.}
\label{fig:local_methods}
\end{figure}

In recent years, biologically plausible models such as spiking neural networks (SNNs) have got attention as energy-efficient alternatives to conventional DNN, owing to their unique spatio-temporal processing capabilities, event-driven nature, and binary spiking activations \cite{Roy2019, Christensen20222022Engineering, Eshraghian2023TrainingLearning}. 
While these features make SNNs promising candidates for enabling energy-efficient inference at the edge, new solutions for solving both the spatial and temporal credit assignment problems are needed for resource-constrained scenarios. 
For example, in an SNN with $L$ layers and $n$ neurons per layer, the backpropagation through time (BPTT) algorithm (Fig.~\ref{fig:local_methods}a) exhibits time and memory complexities of $O(TLn^2)$ and $O(TLn)$, respectively, where $T$ denotes the length of the input sequence. 
This dependency on $T$ makes BPTT impractical for on-device learning on low-power edge devices \cite{Eshraghian2023TrainingLearning, Bellec2020ANeurons, Bohnstingl2022OnlineNetworks}.

To address these limitations, several alternatives have been proposed to achieve local credit assignment. 
For \textit{spatial credit assignment}, methods such as feedback alignment (FA) and direct feedback alignment (DFA) employ random matrices to propagate error signals or directly project errors to individual layers, thereby reducing layer dependencies \cite{Lillicrap2016RandomLearning, Trondheim2016DirectNetworks, Crafton2019DirectLearning}. 
Similarly, the direct random target projection (DRTP) method \cite{Frenkel2021LearningNetworks} projects targets generated from classification labels instead of output errors, allowing for independent and forward layer updates. 
Other biologically plausible approaches replace the backward pass in BP with an additional forward pass \cite{Dellaferrera2022Error-drivenPass, Hinton2022TheInvestigations}. 
Although these methods show promise, they often suffer from slow convergence and scalability issues when applied to deep networks.
For instance, DFA experiences a $\sim16~\%$ accuracy drop compared to BP on CIFAR10 in a five-layer model \cite{Apolinario2024LLS:Synchronization}, while the method proposed in \cite{Dellaferrera2022Error-drivenPass} does not scale beyond two-layer models.
Recently, \cite{Apolinario2024LLS:Synchronization} introduced a local learning rule inspired by neural activity synchronization (LLS), which uses fixed periodic basis vectors for layer-wise training, demonstrating performance comparable to BP in fairly large datasets, including CIFAR100, and Tiny ImageNet.

For \textit{temporal credit assignment} in SNNs, methods inspired by three-factor learning rules leverage eligibility traces \cite{Gerstner2018EligibilityRules} to preserve temporal information in a way that is biologically plausible. 
For example, \cite{Bellec2020ANeurons} introduced e-prop, a method that uses eligibility traces to address temporal credit assignment in SNNs with explicit recurrent connections, while relying on BP or DFA for spatial credit assignment.
Other approaches, such as \cite{Quintana2024ETLP:Hardware, Ortner2023OnlineProjection}, combine eligibility traces with DRTP for spatial credit assignment, achieving spatial and temporal locality. 
However, these methods have proven effective only for shallow models ($2-3$ layers) and scale poorly to deeper architectures due to their high memory cost of $O(Ln^2)$, which, while independent of the number of time steps, scales with the number of synapses ($n^2$). 
An alternative method, the spike-timing-dependent plasticity (STDP)-inspired temporal local learning rule (S-TLLR) \cite{Apolinario2025S-TLLR:Networks}, achieves performance comparable to BPTT with $5-50\times$ less memory cost and $1.3-6.6\times$ less multiply-accumulate (MAC) operations. 
Specifically, S-TLLR memory requirements scale linearly with the number of neurons instead of synapses and independent of the number of time steps, that is $O(Ln)$.
However, S-TLLR still relies on backpropagating errors across layers for spatial credit assignment, as shown for temporally local learning rules in Fig.~\ref{fig:local_methods}b.

To overcome these limitations and enable efficient on-device learning, we propose TESS, a novel scalable temporally and spatially local learning rule for training SNNs (Fig.~\ref{fig:local_methods}c). 
Unlike prior works, which either suffer from scalability issues (e.g., \cite{Bellec2020ANeurons, Quintana2024ETLP:Hardware, Ortner2023OnlineProjection}) or rely on global error propagation across layers (e.g., \cite{Xiao2022OnlineNetworks, Apolinario2025S-TLLR:Networks}), TESS fundamentally addresses the bottlenecks of memory and computational overhead in a way that is both temporally and spatially local. 
Specifically, TESS addresses the temporal credit assignment problem using eligibility traces with memory complexity that scales linearly with the number of neurons, drawing inspiration from \cite{Apolinario2025S-TLLR:Networks}.
For the spatial credit assignment problem, TESS employs a mechanism that synchronizes the activity of neurons within each layer by modulating eligibility traces with a locally generated learning signal, derived from fixed basis vectors inspired by \cite{Apolinario2024LLS:Synchronization}.
This entirely local approach eliminates the need for backpropagation across layers, unlocking scalability to deeper architectures and larger datasets while maintaining memory and computational complexity that are independent of the number of time steps, that is $O(Ln)$ and $O(LCn)$, respectively.

Crucially, TESS marks a significant advancement over prior works by introducing, for the first time, a low-complexity fully local training method for SNNs that achieves performance comparable to BPTT.
On challenging datasets such as CIFAR10-DVS, TESS incurs only a $\sim1.4\%$ accuracy drop relative to BPTT, while on other datasets, including IBM DVS Gesture, and temporal version of CIFAR10, and CIFAR100, TESS matches the performance of BPTT. 
Notably, this is achieved while drastically reducing both computational and memory requirements, with $205-661\times$ fewer MACs and $3-10\times$ lower memory usage.
This capability is achieved through the integration of biologically plausible mechanisms like eligibility traces, STDP, and neural activity synchronization, all of which rely solely on locally available signals within each neuron.

The main contributions of this work are summarized as follows:  
\begin{itemize}  
\item We propose TESS, a novel scalable learning rule for SNNs that integrates biologically inspired mechanisms such as eligibility traces, STDP, and neural activity synchronization. TESS operates in a fully local manner, relying only on locally available signals, making it well-suited for energy-efficient on-device learning.  

\item TESS achieves linear memory complexity, $O(Ln)$, and computational complexity, $O(LCn)$, enabling the efficient training of deeper architectures, such as VGG-9, on resource-constrained edge devices.  

\item TESS delivers performance comparable to BPTT on vision benchmarks, matching its accuracy on IBM DVS Gesture, CIFAR10, and CIFAR100, while incurring only a $\sim1.4\%$ accuracy drop on CIFAR10-DVS. This is achieved with significantly lower resource requirements.  
\end{itemize}

\section{Background}\label{sec:background}
In this section, we introduce the spiking neuron models adopted here, the mathematical notation, gradient-based optimization approaches for SNNs, and three factor learning rules. 
        \subsection{LIF model}\label{sec:lif}
        We adopt the leaky integrate-and-fire (LIF) neuron model to implement spiking behavior. 
        The discrete LIF neuron model is mathematically described as follows:
        \begin{equation}
            \label{eq:lif_integrate}
            \vu^{(l)}_{i}[t] = \gamma (\vu^{(l)}_{i}[t-1] - v_{\text{th}}\vo^{(l)}_{i}[t-1]) + \sum_j\mW^{(l)}_{ij}\vo^{(l-1)}_{j}[t],
        \end{equation}
        \begin{equation}
            \label{eq:lif_fire}
            \vo^{(l)}_{i}[t] = \Theta(\vu^{(l)}_{i}[t]-v_{\text{th}}),
        \end{equation}
        where, $\vu^{(l)}_i[t]$ represents the membrane potential of the $i$-th neuron in layer $l$ at the time step $t$, and $\mW^{(l)}_{ij}$ is the strength of the synaptic connection between the $i$-th post-synaptic neuron in layer $l$ and the $j$-th pre-synaptic neuron in layer $l-1$.
        The parameter $\gamma$ is the leak factor, producing an exponential decay of the membrane potential over time.
        The threshold voltage is denoted by $v_{\text{th}}$, and $\Theta$ represents the Heaviside function ($\Theta(x) = 1$ if $x > 0$ and $0$ otherwise).
        When $\vu^{(l)}_i[t]$ reaches $v_{\text{th}}$, the neuron fires, producing a binary spike output $\vo^{(l)}_i[t] = 1$. 
        This firing triggers a subtractive reset mechanism, represented by the reset signal $v_{\text{th}}\vo^{(l)}_{i}[t]$, which reduces $\vu^{(l)}_i[t]$ by $v_{\text{th}}$.

        \subsection{Gradient-based optimization for SNNs}\label{sec:bptt}
        We now describe how gradient-based optimization is applied to SNNs.

        Given a dataset $\sD = \{(\vx, \vy^*)_{i=1}^N\}$, where $N$ is the number of samples, $\vx$ represents the input data, and $\vy^*$ denotes the corresponding labels, and an SNN model with parameters $\sW = \{ \mW^{(l)} \}_{l=1}^L$, where $L$ is the number of layers, the optimization goal is to minimize a loss function $\mathcal{L}$,
        $$
            \sW := \arg\min_{\sW} \mathcal{L}(\sD; \sW).
        $$
        This minimization is solved using (stochastic) gradient descent, where the parameters are iteratively updated as:
        $$
            \mW^{(l)} := \mW^{(l)} - \eta \frac{d\mathcal{L}}{d\mW^{(l)}},
        $$
        where $\eta$ is the learning rate and $\frac{d\mathcal{L}}{d\mW^{(l)}}$ represents the gradient of the loss function with respect to the parameters of the $l$-th layer.
        The gradients are computed using the BPTT algorithm, which applies the chain rule over both space (i.e.~layers) and time:
        $$
            \frac{d\mathcal{L}}{d\mW^{(l)}} = \sum_{t=1}^T \frac{\partial \mathcal{L}}{\partial \vu^{(l)}[t]} \frac{\partial \vu^{(l)}[t]}{\partial \mW^{(l)}},
        $$
        where $T$ is the total number of time-steps of the input sequence $\vx$. 
        Due to the recurrent nature of SNNs, $\frac{\partial \mathcal{L}}{\partial \vu^{(l)}[t]}$ depends on the entire history of the model:
        $$
            \frac{\partial\mathcal{L}}{\partial\vu^{(l)}[t]}=\frac{\partial\mathcal{L}}{\partial\vo^{(l)}[t]}\frac{\partial\vo^{(l)}[t]}{\partial\vu^{(l)}[t]}+\frac{\partial\mathcal{L}}{\partial\vu^{(l)}[t+1]}\frac{\partial\vu^{(l)}[t+1]}{\partial\vu^{(l)}[t]},
        $$
        where the term $\frac{\partial \mathcal{L}}{\partial \vo^{(l)}[t]}$ requires information from all subsequent layers ($L-l$), while $\frac{\partial \mathcal{L}}{\partial \vu^{(l)}[t+1]} \frac{\partial \vu^{(l)}[t+1]}{\partial \vu^{(l)}[t]}$ depends on the full temporal history. 
        Thus, BPTT is neither spatially nor temporally local and incurs high computational costs.  

    \subsection{Three-factor learning rules}\label{sec:3f_lr}
    Three-factor learning rules \cite{Gerstner2018EligibilityRules} represent a biologically plausible framework for synaptic plasticity, where synapse updates are determined by the interaction of three key factors: pre-synaptic activity, post-synaptic activity, and a top-down learning signal. 

    The core idea of three-factor learning rules is the concept of an eligibility trace ($\ve_{ij}$), which is a transient variable that encodes synaptic changes driven by pre- and post-synaptic activity. 
    This trace persists over time, allowing updates to occur when a delayed top-down learning signal arrives. 
    The eligibility trace is typically modeled as a function of the pre- and post-synaptic activity, decaying over time according to the following recurrent formulation:
    \begin{equation}
        \ve^{(l)}_{ij}[t] = \beta \ve^{(l)}_{ij}[t-1] + f(\vo^{(l)}_i[t])g(\vo^{(l-1)}_j[t]),
        \label{eq:eligibility_trace}
    \end{equation}
    where $\beta$ is an exponential decay factor, $f(\vo^{(l)}_i[t])$ and $g(\vo^{(l-1)}_j[t])$ are element-wise functions of the post-synaptic activity $\vo^{(l)}_i[t]$ and pre-synaptic activity $\vo^{(l-1)}_j[t]$, respectively. 
    This formulation ensures that synapses are ``eligible'' for updates only when neuronal activity meets certain conditions.
    
    The actual synaptic update is computed by modulating the eligibility trace with a top-down learning signal $\vm_i[t]$, which represents error or reward information. 
    The weight update rule can be expressed as:
    \begin{equation}
        \Delta \mW_{ij} = \sum_{t} \vm_i[t] \ve^{(l)}_{ij}[t],
    \end{equation} 
    where the learning signal $\vm_i[t]$ is typically derived from task-relevant objectives, such as the gradient of a loss function or a reward signal. 
    This modulation ensures that synaptic updates are oriented towards minimizing a particular objective.
    
    Three-factor learning rules have demonstrated effectiveness in training artificial SNNs, as evidenced by \cite{Bellec2020ANeurons, Kaiser2020SynapticDECOLLE, Xiao2022OnlineNetworks, Ortner2023OnlineProjection, Apolinario2025S-TLLR:Networks, Bohnstingl2022OnlineNetworks}. 
    Notably, they offer a biologically plausible approximation of BPTT under specific conditions \cite{Bellec2020ANeurons, Martin-Sanchez2022ARules}.   
    Despite their promise, previous works leveraging eligibility traces for temporal credit assignment still rely on global error propagation across layers, Fig.~\ref{fig:local_methods}b, for achieving performance comparable to BPTT \cite{Bohnstingl2022OnlineNetworks, Xiao2022OnlineNetworks, Apolinario2025S-TLLR:Networks}.
    Which results in a time complexity of $O(Ln^2)$.

\section{Proposed Method - A Scalable  Fully Local Learning Rule}\label{sec:method}

\begin{figure}[t]
\centering
\includegraphics[width=\columnwidth]{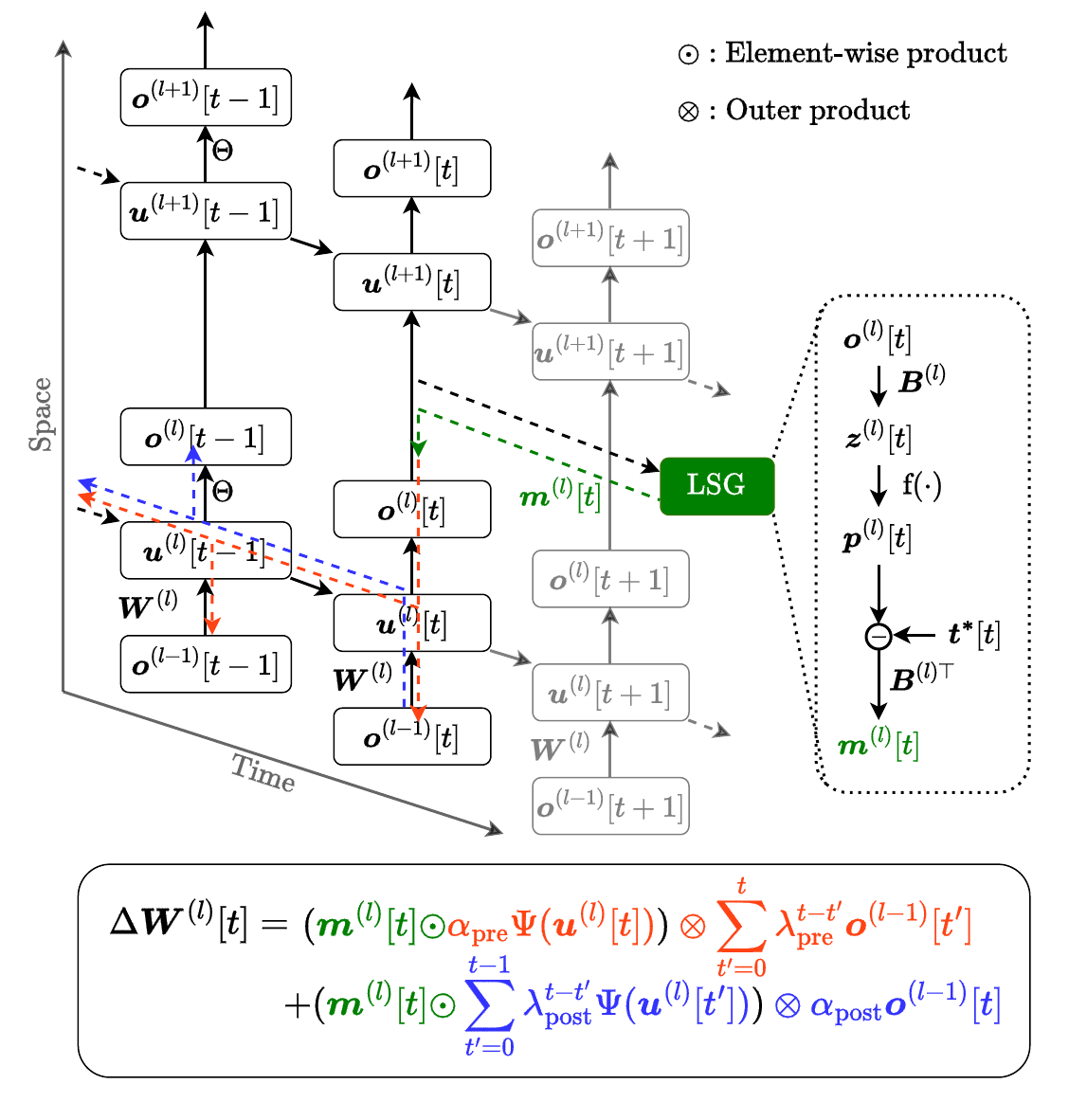}
\caption{Overview of TESS. The diagram illustrates an SNN model unrolled in time, where $\vu^{(l)}[t]$ denotes the membrane potential of neurons in the $l$-th layer at time step $t$, and $\vo^{(l)}[t]$ represents the corresponding output spikes. Signals involved in weight update computation are highlighted: {\color{myred}red} represents the eligibility trace based on causal relationships between inputs and outputs, {\color{myblue}blue} represents the eligibility trace for non-causal relationships, and {\color{mygreen}green} represents the local learning signal $\vm^{(l)}[t]$ used to modulate the eligibility traces. The local learning signal is generated independently for each layer through a learning signal generation (LSG) process. The fixed binary matrix $\mB^{(l)}$ used in the LSG process features columns corresponding to square wave functions. While, $f(\cdot)$ is a softmax function, and $\vt^*[t]$ represent the labels.} 
\label{fig:method_diagram}
\end{figure}

To address the challenges of temporal and spatial credit assignment for SNNs, we propose TESS, a scalable temporally and spatially local learning rule that is biologically inspired. 
It is designed as a three-factor learning rule that operates efficiently with low computational and memory overhead, making it suitable for resource-constrained edge devices.
Specifically, TESS is presented in Fig.~\ref{fig:method_diagram} and relies on two key components to achieve temporal and spatial locality:

\paragraph{Temporal Credit Assignment with Eligibility Traces}
As discussed in Section~\ref{sec:3f_lr}, eAs discussed in Section~\ref{sec:3f_lr}, eligibility traces are transient variables driven by changes in pre- and post-synaptic activity. 
These traces encode the temporal history of synaptic connections, identifying synapses as candidates for updates and thereby addressing the temporal credit assignment problem by tracking neuronal activity.  

However, eligibility traces as formulated in (\ref{eq:eligibility_trace}) incur a memory complexity of $O(n^2)$, which is prohibitive for deep SNN models. 
To overcome this, and in line with prior works \cite{Frenkel2022ReckOn:Timescales, Xiao2022OnlineNetworks, Apolinario2025S-TLLR:Networks}, we restrict the formulation to instantaneous eligibility traces by setting $\beta=0$ in (\ref{eq:eligibility_trace}). 
This modification reduces the memory complexity to $O(n)$ by independently tracking pre- and post-synaptic activity traces.  

In TESS, we utilize two eligibility traces: one based on pre-synaptic activity (shown in {\color{myred}red} in Fig.~\ref{fig:method_diagram}) and one based on post-synaptic activity (shown in {\color{myblue}blue} in Fig.~\ref{fig:method_diagram}). 
These two components mimic STDP mechanisms, capturing causal ({\color{myred}red signals}) and non-causal ({\color{myblue}blue signals}) dependencies between pre- and post-synaptic activity.  

We first consider the eligibility trace with causal information. Using (\ref{eq:eligibility_trace}), the function $f(\cdot)$ is defined as a secondary activation function $\Psi(\cdot)$ applied to the membrane potential $\vu^{(l)}[t]$. 
This serves a role analogous to surrogate gradients in gradient-based approaches \cite{Neftci2019SurrogateNetworks}. 
The function $g(\cdot)$ is a low-pass filter applied to the input spikes:  
$
{\color{myred} \sum_{t'=0}^{t} \lambda_{\text{pre}}^{t-t'} \vo^{(l-1)}[t']},
$  
where $\lambda_{\text{pre}}$ is an exponential decay factor, representing the trace of pre-synaptic activity. 
To compute this trace in a forward-in-time manner, we introduce a recurrent variable $\vq^{(l)}[t]$, defined as:  
\begin{equation}  
{\color{myred}  
\vq^{(l)}[t] = \lambda_{\text{pre}} \vq^{(l)}[t-1] + \vo^{(l-1)}[t]},  
\label{eq:input_trace}  
\end{equation}  
which allows the causal eligibility trace to be expressed as:  
\begin{equation}  
{\color{myred}  
\ve^{(l)}_{\text{pre}}[t] = \alpha_{\text{pre}} \Psi(\vu^{(l)}[t]) \otimes \vq^{(l)}[t]},  
\label{eq:e_trace_pre}  
\end{equation}  
where $\alpha_{\text{pre}}$ controls the amplitude of the eligibility trace. 
For all experiments, $\alpha_{\text{pre}}$ is set to 1.  

For the second eligibility trace, ${\color{myblue} \ve^{(l)}_{\text{post}}[t]}$, we use a low-pass filter over the activations of the membrane potential $\Psi(\vu^{(l)}[t])$:  
$
{\color{myblue} \sum_{t'=0}^{t-1} \lambda_{\text{post}}^{t-t'} \Psi(\vu^{(l)}[t'])}.
$  
This can also be expressed as a recurrent equation by introducing a new variable $\vh^{(l)}[t]$:  
\begin{equation}  
{\color{myblue}  
\vh^{(l)}[t] = \lambda_{\text{post}} \vh^{(l)}[t-1] + \Psi(\vu^{(l)}[t-1])},  
\label{eq:output_trace}  
\end{equation}  
and the non-causal eligibility trace is then given by:  
\begin{equation}  
{\color{myblue}  
\ve^{(l)}_{\text{post}}[t] = \alpha_{\text{post}} \vh^{(l)}[t] \otimes \vo^{(l-1)}[t]},  
\label{eq:e_trace_post}  
\end{equation}  
where $\alpha_{\text{post}}$ determines the inclusion of non-causal terms, with $\alpha_{\text{post}} = +1$ for positive inclusion, $\alpha_{\text{post}} = -1$ for negative inclusion, and $\alpha_{\text{post}} = 0$ for exclusion.

\paragraph{Spatial Credit Assignment with Locally Generated Learning Signals}\label{sec:learning_signal}
As discussed in Section~\ref{sec:3f_lr}, while eligibility traces address the temporal credit assignment, synaptic updates require a top-down learning signal, denoted as $\color{mygreen} \vm^{(l)}[t]$, to modulate the eligibility traces and solve the spatial credit assignment problem. 
Unlike prior approaches that rely on the global backpropagation of errors to compute this learning signal, TESS introduces a local mechanism for spatial credit assignment through a {\color{mygreen}learning signal generation (LSG)} process. 
This method, depicted in Fig.~\ref{fig:method_diagram}, enables the learning signals to be computed locally within each layer, avoiding the computational overhead of global error propagation and making the approach both scalable and hardware-friendly.  

The LSG process begins by projecting the output spikes of each layer, $\vo^{(l)}[t]$, into a $C$-dimensional task subspace using a projection matrix $\mB^{(l)}$. 
This projection captures the relevant task information for the layer. 
Next, a function $f(\cdot)$ is applied to the projected vector to compute its alignment with the target vector $\vy^*$. 
The choice of $f(\cdot)$ depends on the task: for classification problems, $f(\cdot)$ is the softmax function, while for regression tasks, it is simply the identity function. 
The difference between the projected vector and the target, $f(\mB^{(l)}\vo^{(l)}[t]) - \vy^*$, serves as the error signal. 
This error signal is then projected back to the layer using the transpose of the projection matrix, $\mB^{(l)\top}$, resulting in the modulatory learning signal, $\vm^{(l)}[t]$, which is given by:  
\begin{equation}  
{\color{mygreen}  
\vm^{(l)}[t] = \mB^{(l)\top} \left( f(\mB^{(l)} \vo^{(l)}[t]) - \vy^* \right)  
} . 
\label{eq:learning_signal}  
\end{equation}  

The design of the projection matrix $\mB^{(l)}$ is a critical component of the LSG process. 
In TESS, $\mB^{(l)}$ is defined as a fixed binary matrix, with each column corresponding to a square wave function. 
This design offers several advantages. 
The square wave functions help synchronize the activity of neurons within the same layer by assigning distinct spatial frequencies to different classes, ensuring that the task-related information is distributed effectively across the layer. 
Furthermore, the columns of $\mB^{(l)}$ are quasi-orthogonal, minimizing interference between the projections of different classes. 
The simplicity of square wave functions also makes them highly hardware-efficient, which is particularly advantageous for implementation in resource-constrained environments.  

The locally generated learning signal in TESS eliminates the need for global backpropagation, significantly reducing computational complexity from $O(n^2)$ to $O(Cn)$. 
The effectiveness of this design has been demonstrated in prior work \cite{Apolinario2024LLS:Synchronization}, further validating its potential for real-world applications.

\paragraph{Weight Updates}
The weight updates are computed based on the interaction of the learning signals with the eligibility traces. The updates for causal terms ($\mW^{(l)}_{\text{pre}}$) and non-causal terms ($\mW^{(l)}_{\text{post}}$) contributions are given by:
\begin{equation}
    \Delta \mW^{(l)}_{\text{pre}}[t] = \left({ \color{mygreen}\vm^{(l)}[t]\odot}  \color{myred}\alpha_{\text{pre}} \Psi(\vu^{(l)}[t])\right){\color{myred} \otimes\vq^{(l)}[t] },
    \label{eq:weight_update_pre}
\end{equation}
\begin{equation}
    \Delta \mW^{(l)}_{\text{post}}[t] =  \left({ \color{mygreen}\vm^{(l)}[t]\odot}  \color{myblue}\alpha_{\text{post}} \vh^{(l)}[t]\right){\color{myblue} \otimes\vo^{(l-1)}[t]}.
    \label{eq:weight_update_post}
\end{equation}

The total weight update combines these two contributions:
\begin{equation}
    \Delta \mW^{(l)}[t]=\Delta \mW^{(l)}_{\text{pre}}[t] + \Delta \mW^{(l)}_{\text{post}}[t]
    \label{eq:weight_update_total}
\end{equation}

\begin{algorithm}[t]
    \caption{TESS pseudo-code for layer $l$}\label{alg:pseudocode}
    \begin{algorithmic}
    \Require $\vo^{(l-1)}$ (input for layer $l$), $\mB$ (fixed binary matrix), $\beta$ (threshold), $\eta$ (learning rate), $t_l$ (time step to start generating the learning signal)
    \Ensure $\mW^{(l)}$
    \State $\vu^{(l)}[0]=0$
    \State $\vh^{(l)}[0]=0$
    \State $\vq^{(l)}[0]=0$
    \For {$t=1, 2, \dots, T$}
        \State Update $\vh^{(l)}[t]$ based on (\ref{eq:output_trace})
        \State Update $\vu^{(l)}[t]$ and $\vo^{(l)}[t]$ based on (\ref{eq:lif_integrate}) and (\ref{eq:lif_fire})
        \State Update $\vq^{(l)}[t]$ based on (\ref{eq:input_trace})
    
        \If {$t >= t_l$} 
            \State Compute $\vm^{(l)}[t]$ based on (\ref{eq:learning_signal})
            \State Compute $\Delta\mW^{(l)}[t]$ based on (\ref{eq:weight_update_pre}), (\ref{eq:weight_update_post}), and (\ref{eq:weight_update_total})
        \EndIf
        
    \EndFor
    \State $\mW^{(l)}=\mW^{(l)}+\eta \sum_{t=t_l}^{T}\Delta\mW^{(l)}[t]$
    
    \end{algorithmic}
    \end{algorithm}

    \subsection{Algorithm Implementation}
    The TESS algorithm operates iteratively, updating eligibility traces, computing learning signals, and adjusting weights for each time step. 
    A pseudo-code implementation for layer $l$ is provided in Algorithm~\ref{alg:pseudocode}.

    \subsection{Computational and Memory Cost}\label{sec:computational_cost}
    In this subsection, we analyze the theoretical computational improvements of TESS in terms of multiply-accumulate (MAC) operations and memory requirements, comparing it to BPTT and S-TLLR. 
    We build on the analysis presented in \cite{Apolinario2025S-TLLR:Networks}, which we expanded to include the effects of the spatial and temporal locality of TESS on computational and memory costs.

    \subsubsection{Memory Requirements}\label{sec:mem_cost}
    We begin by discussing memory requirements, focusing on the overhead associated with synaptic updates, excluding the state variables required for SNN inference (e.g., membrane potential)
    According to \cite{Apolinario2025S-TLLR:Networks}, the memory requirements for BPTT and S-TLLR can be estimated using the following equations:
    \begin{equation}
    \text{Mem}_{\text{BPTT}} = T \sum_{l=0}^L n^{(l)}\text{ , }
    \text{Mem}_{\text{S-TLLR}} = 2  \sum_{l=0}^L n^{(l)},
    \label{eq:mem_stllr}
    \end{equation}
    where, $n^{(l)}$ represents the number of neurons in layer $l$, $L$ is the total number of layers in the model, and $T$ is the total number of time steps (length of the input sequence). 
    The factor of 2 in (\ref{eq:mem_stllr}) arises from the inclusion of both causal and non-causal terms in the computation of eligibility traces when $\alpha_{\text{pre}}$ and $\alpha_{\text{post}}$ are nonzero.

    For TESS, the memory requirements are determined by analyzing the variables involved in  (\ref{eq:weight_update_pre}) and (\ref{eq:weight_update_post}).
    From (\ref{eq:weight_update_pre}), $\vm^{(l)}[t]$ depends on the output spikes $\vo^{(l)}[t]$, which are computed using (\ref{eq:learning_signal}). Since this signal is derived directly from the current output spikes, it does not require additional memory storage and can be computed on the fly.
    Similarly, $\Psi(\vu^{(l)}[t])$ is a function of the current membrane potential and does not require additional memory, as it can also be computed on the fly.
    In contrast, the term $\vq^{(l)}[t]$ accounts for the history of pre-synaptic activity and requires memory proportional to the number of input neurons, $n^{(l-1)}$. 
    Likewise, $\vh^{(l)}[t]$ represents the history of post-synaptic activity and requires memory proportional to the number of output neurons, $n^{(l)}$.
    By combining these terms, the total memory requirement for TESS can be expressed as:
    \begin{equation}
        \label{eq:tess_mem}
         \text{Mem}_{\text{TESS}} = 2  \sum_{l=0}^L n^{(l)}.
    \end{equation}
    This demonstrates that TESS achieves memory efficiency comparable to S-TLLR while avoiding the significant overhead associated with the time-dependent storage of BPTT.
    
    \subsubsection{Computational Requirements}\label{sec:computation_req}
    Here, we estimate the computational requirements by evaluating the number of MAC operations needed to compute the learning signals. 
    Specifically, we compare the operations required to calculate $\frac{\partial\mathcal{L}}{\partial\vu^{(l)}[t]}$ for BPTT, $[\frac{\partial\mathcal{L}}{\partial\vu^{(l)}[t]}]_{\text{local}}$ for S-TLLR, and $\vm^{(l)}[t]$ for TESS. 
    For simplicity, we assume a fully connected network and disregard any element-wise operations.  

    For both BPTT and S-TLLR, the error signals are computed by propagating errors from the last layer to the first. 
    If the final error vector has a dimension of $n^{(L)}$, it is propagated to the previous layer using the weight matrix $\mW^{(L)}$, which has a dimension of $n^{(L)} \times n^{(L-1)}$. 
    This matrix-vector multiplication requires $n^{(L)} \times n^{(L-1)}$ MAC operations. The same process is repeated for all hidden layers.  
    However, while BPTT performs this backpropagation for all time steps $T$, S-TLLR computes the learning signal only for the final $T - t_l$ time steps. 
    Thus, the number of MAC operations can be expressed as: 
    \begin{equation}
    \text{MAC}_{\text{BPTT}} = T \sum_{l=1}^L n^{(l)} \times n^{(l-1)},
    \label{eq:bptt_mac}
    \end{equation}
    \begin{equation}
    \text{MAC}_{\text{S-TLLR}} = (T-t_l) \sum_{l=1}^L n^{(l)} \times n^{(l-1)}.
    \label{eq:stllr_mac}
    \end{equation}

    In contrast to BPTT and S-TLLR, TESS generates the learning signal locally for each layer, significantly reducing the computational complexity associated with backpropagating errors through layers and time. 
    The TESS learning signal is computed using (\ref{eq:learning_signal}), where the output spikes $\vo^{(l)}[t]$ (of dimension $n^{(l)}$) are projected into a task subspace of dimension $C$ using a binary fixed matrix $\mB^{(l)}$ of dimension $C \times n^{(l)}$.
    This projection requires $2 \times C \times n^{(l)}$ MAC operations to compute $\vm^{(l)}$. 
    Therefore, the total number of MAC operations for a network with $L$ layers is: 
    \begin{equation}
    \text{MAC}_{\text{TESS}} = (T-t_l)  \sum_{l=1}^{L} 2 \times n^{(l)} \times C
    \label{eq:tess_mac}
    \end{equation}
    where $C$ represents the number of classes in a classification task or the number of variables to estimate in a regression task. 
    By generating the learning signal locally, TESS achieves a significant reduction in computational complexity compared to BPTT and S-TLLR. 
    Specifically, the reduction factor is approximately $\frac{n^{(l)}}{C}$, as $C \ll n^{(l)}$ in most practical scenarios.

    \begin{table}[t]
    \caption{Comparison of TESS with other learning rules for training spiking neural networks (SNNs). The parameters are defined as follows: $L$ represents the number of layers, $n$ is the average number of neurons per layer, $T$ is the total number of time steps, and $C$ denotes the number of targets.}
    \label{table:method_comparison}
    \centering
    \resizebox{\columnwidth}{!}{
    \begin{tabular}{lcccc}
    \toprule
    \multicolumn{1}{c}{\textbf{Method}} & \textbf{\begin{tabular}[c]{@{}c@{}}Memory \\ Complexity\end{tabular}} & \textbf{\begin{tabular}[c]{@{}c@{}}Time\\ Complexity\end{tabular}} & \textbf{\begin{tabular}[c]{@{}c@{}}Temporal \\ Locality\end{tabular}} & \textbf{\begin{tabular}[c]{@{}c@{}}Spatial \\  Locality \end{tabular}} \\ 
    \midrule
    BPTT & $TLn$ & $TLn^2$ & \color{red}\xmark & \color{red}\xmark \\ 
    e-prop \cite{Bellec2020ANeurons} & $Ln^2$ & $Ln^2$ & \color{green}\cmark & \color{red}\xmark \\ 
    OSTL \cite{Bohnstingl2022OnlineNetworks}& $Ln^2$  & $Ln^2$ & \color{green}\cmark & \color{red}\xmark \\ 
    ETLP \cite{Quintana2024ETLP:Hardware} & $Ln^2$  & $LCn$ & \color{green}\cmark & \color{green}\cmark  \\ 
    OSTTP \cite{Ortner2023OnlineProjection} & $Ln^2$   & $LCn$& \color{green}\cmark & \color{green}\cmark \\ 
    OTTT \cite{Xiao2022OnlineNetworks} & $Ln$  & $Ln^2$ & \color{green}\cmark & \color{red}\xmark \\ 
    S-TLLR \cite{Apolinario2025S-TLLR:Networks} & $Ln$  & $Ln^2$ & \color{green}\cmark & \color{red}\xmark \\
    \textbf{TESS (Ours)} & $Ln$  & $LCn$ & \color{green}\cmark & \color{green}\cmark \\ 
    \bottomrule
    \end{tabular}}
    \end{table}

    \subsection{Comparison with other local learning rules}\label{sec:comparison_llr}
    In this subsection, we analyze the time and memory complexity of TESS in comparison to other approaches. 
    For this analysis, we consider a fully connected spiking neural network with $L$ layers, each containing $n$ neurons, trained on temporal tasks with $T$ time steps, and a target space of dimensionality $C$. 

    As discussed in Section~\ref{sec:bptt}, BPTT requires access to the entire history of the network, resulting in a memory complexity of $O(TLn)$. 
    Similarly, since learning signals are produced by propagating errors through layers for all time steps, the time complexity is $O(TLn^2)$. This implies that tasks with greater temporal dependencies significantly increase the cost of BPTT.

    To address this dependency on time steps, temporal local methods such as e-prop \cite{Bellec2020ANeurons}, OSTL \cite{Bohnstingl2022OnlineNetworks}, OTTT \cite{Xiao2022OnlineNetworks}, and S-TLLR \cite{Apolinario2025S-TLLR:Networks}, as well as fully local methods like OSTTP \cite{Ortner2023OnlineProjection} and ETLP \cite{Quintana2024ETLP:Hardware}, leverage eligibility traces. 
    This strategy allows them to maintain a memory requirement independent of time steps. 
    However, methods like e-prop, OSTL, ETLP, and OSTTP exhibit a memory complexity of $O(Ln^2)$, which can become prohibitively expensive for large models. 
    In contrast, methods such as OTTT and S-TLLR achieve a more efficient linear memory complexity of $O(Ln)$, making them more scalable. 
    Similarly, TESS has been designed to exhibit linear memory complexity.

    Regarding time complexity, methods such as e-prop, OSTL, OTTT, and S-TLLR rely on backpropagation of errors through layers to generate learning signals, incurring a complexity of $O(Ln^2)$. 
    In contrast, OSTTP and ETLP use the DRTP mechanism \cite{Frenkel2021LearningNetworks} to achieve spatial locality, reducing the time complexity to $O(LCn)$, where $C \ll n$. 
    TESS follows a similar approach, generating learning signals locally and achieving the same reduced time complexity of $O(LCn)$.

    Compared to other methods, TESS offers the best combination of memory and time complexity due to its spatial and temporal locality features. 
    Furthermore, TESS achieves this efficiency while delivering performance comparable to methods with higher memory and time requirements, as discussed in the next section. 
    A summary of this comparison is presented in Table~\ref{table:method_comparison}.

\section{Experimental Evaluation}
In this section, we evaluate the performance of our training algorithm, TESS, on multiple datasets, assessing its ability to achieve competitive accuracy at reduced cost.
To do so, we compare TESS with a broad range of non-local to local learning state-of-the-art methods.

    \subsection{Experimental Setup}
    
    This subsection describes the datasets, pre-processing steps, and model architectures used to evaluate our method.

        \subsubsection{Datasets}
        \label{sec:datasets}
        We evaluated TESS using four datasets: \mbox{CIFAR10} \cite{Krizhevsky2009LearningImages}, CIFAR100 \cite{Krizhevsky2009LearningImages}, IBM DVS Gesture \cite{Amir2017ASystem}, and CIFAR10-DVS \cite{Li2017CIFAR10-DVS:Classification}. 
        The preprocessing steps for each dataset are as follows:
        \begin{itemize}
            \item CIFAR10 and CIFAR100: Images were presented to the SNN models for 6 time steps to simulate a temporal dimension. 
            Data augmentation during training included increasing image size via zero-padding of 4, random cropping to $32\times32$, application of the cutout technique \cite{Devries2017ImprovedCutout}, random horizontal flipping, and normalization.
            \item CIFAR10-DVS: Events were accumulated in 10 frame events and resized to $48\times48$. 
            Data augmentation involved random cropping with zero-padding of 4, followed by normalization.
            \item IBM DVS Gesture: Sequences of varying lengths were split into samples of $1.5$ seconds. 
            Events were accumulated into 20 event frames, each representing $75$ ms, resized to $32\times32$, and randomly cropped with zero-padding of 4.
        \end{itemize}

        \subsubsection{Training details}
        All experiments were conducted on a VGG-9 model using the Adam optimizer with a learning rate of $0.001$, and models were trained for 200 epochs. 
        A learning rate scheduler was employed to reduce the learning rate by half if the validation accuracy did not improve for 5 consecutive epochs. 
        The exponential decay factors, $\lambda_{\text{pre}}$ and $\lambda_{\text{post}}$, were set to $0.5$ and $0.2$, respectively, while $\alpha_{\text{pre}}$ was fixed at $1$.  
        The leak factor ($\gamma$) and the threshold ($v_{\text{th}}$) of the LIF model were set to $0.5$ and $0.6$, respectively. 
        Additionally, the secondary activation function was chosen to be a triangular function, defined as $\Psi(\vu) = 0.3 \cdot \max(1 - |\vu - v_{\text{th}}|, 0)$. 
        Finally, weight updates occurred at every time step, i.e. $t_l = 0$.

\begin{table}[t]
    \renewcommand{\arraystretch}{1.3}
    \caption{Ablation study on including non-causal terms ($\alpha_{\text{post}}$) in the eligibility traces during learning. Accuracy (mean$\pm$std) reported over five independent trials}
    \label{table:ablation_alpha_post}
    \centering
    \resizebox{0.5\textwidth}{!}{
    \begin{tabular}{lcccc}
    \toprule
    \multicolumn{1}{c}{\textbf{Dataset}} & \multicolumn{1}{c}{$\boldsymbol{T}$} & \multicolumn{1}{c}{\textbf{$\boldsymbol{\alpha_{post}=-1}$}} & \multicolumn{1}{c}{\textbf{$\boldsymbol{\alpha_{post}=0}$}}  & \multicolumn{1}{c}{\textbf{$\boldsymbol{\alpha_{post}=+1}$}}  \\ 
    \midrule
    CIFAR10-DVS & $10$  & $75.00\pm0.69\%$ & $\mathbf{75.00\pm0.65\%}$ & $74.36\pm0.87\%$  \\ 
    DVS Gesture & $20$ & $98.56\pm0.41\%$ & $98.33\pm0.57\%$ & $\mathbf{98.56\pm0.31\%}$  \\
    CIFAR10 & $6$  &  $89.93\pm0.31\%$ & $91.99\pm0.19\%$ & $\mathbf{92.55\pm0.16\%}$  \\
    CIFAR100 & $6$  & $62.49\pm1.05\%$ & $68.19\pm0.55\%$ & $\mathbf{70.00\pm0.34\%}$\\
    \bottomrule
    \end{tabular}
    }
    \end{table}

    \begin{table*}[!h]
    \caption{Comparison of performance and computational requirements of different local and non-local learning strategies on image classification tasks }
    \label{table:performance}
    \centering
    \resizebox{0.8\textwidth}{!}{
    \begin{tabular}{lccccccc}
    \toprule
    \multicolumn{1}{c}{\textbf{Method}} & \multicolumn{1}{c}{\textbf{Model}}& \textbf{\begin{tabular}[c]{@{}c@{}}Local\\ Learning\end{tabular}} &  \textbf{\begin{tabular}[c]{@{}c@{}}Time-steps\\ ($T$)\end{tabular}} & \textbf{\begin{tabular}[c]{@{}c@{}}Batch\\ Size\end{tabular}}  & \textbf{\begin{tabular}[c]{@{}c@{}}Accuracy$^1$ \\ (mean$\pm$std)\end{tabular}} & \textbf{\begin{tabular}[c]{@{}c@{}}\# MAC$^2$\\ ($\times 10^6$)\end{tabular}} & \textbf{\begin{tabular}[c]{@{}c@{}}Memory$^2$\\ (MB)\end{tabular}} \\ 
    \midrule        
    \multicolumn{8}{c}{CIFAR10-DVS} \\ \midrule
    BPTT \cite{Fang2021IncorporatingNetworks} & PLIF (7 layers) & \color{red}No & 20 & 16 & $74.8\%$ & - & -\\ 
    TET \cite{Deng2022TemporalRe-weighting} & VGG-11& \color{red}No & \multicolumn{1}{c}{10} & 128 & $83.17\pm0.15\%$ & - & -\\ 
    DSR \cite{Meng2022TrainingRepresentation} & VGG-11& \color{red}No & 10 & 128 & $77.27\pm0.24\%$ & - & -\\ 
    OTTT\textsubscript{A}\cite{Xiao2022OnlineNetworks} & VGG-9 & Partial (in time) & 10 & 128 & $76.27\pm0.05\%$ & - & -\\ 
    BPTT \cite{Apolinario2025S-TLLR:Networks} & VGG-9& \color{red}No & 10 & 48 & $75.44\pm0.76\%$ & - & - \\ 
    S-TLLR \cite{Apolinario2025S-TLLR:Networks} & VGG-9& Partial (in time) & 10 & 48 & $75.6\pm0.10\%$ & - & -\\
    \textbf{BPTT (baseline)} & VGG-9 & \color{red}No & 10 & 64 & $76.40\pm0.66\%$ & $13589.59$ & $25.50$ \\ 
    \textbf{S-TLLR (baseline)} & VGG-9 & Partial (in time) & 10 & 64 & $75.14\pm1.37\%$ & $13589.59$ & $5.10$\\
    \textbf{TESS (ours)} & VGG-9 & \color{green}Yes & 10 & 64 & $75.00\pm0.65\%$ & $22.15$ & $2.55$\\
    \midrule
    \multicolumn{8}{c}{IBM DVS Gesture } \\ \midrule
    SLAYER \cite{Shrestha2018SLAYER:Time}$^3$ & SNN (8 layers) & \color{red}No & 300 & - & $93.64\pm0.49\%$ & - & -\\ 
    DECOLLE\cite{Kaiser2020SynapticDECOLLE} & SNN (4 layers) & \color{green}Yes & 1800 & 72 & $95.54\pm0.16\%$ & - & -\\ 
    OTTT\textsubscript{A}\cite{Xiao2022OnlineNetworks}$^3$ & VGG-9 & Partial (in time) & 20 & 16 & $96.88\%$ & - & -\\ 
    BPTT \cite{Apolinario2025S-TLLR:Networks} & VGG-9 & \color{red}No & 20 & 16 & $95.58\pm1.08\%$ & - & -\\ 
    S-TLLR \cite{Apolinario2025S-TLLR:Networks} & VGG-9 & Partial (in time) & 20 & 16 & $97.72\pm0.38\%$ & - & -\\ 
    \textbf{BPTT (baseline)} & VGG-9 & \color{red}No & 20 & 16 & $97.95\pm0.68\%$ & $12079.69$ & $22.69$\\ 
    \textbf{S-TLLR (baseline)} & VGG-9 & Partial (in time) & 20 & 16 & $98.48\pm0.37\%$ & $12079.69$ & $2.26$ \\ 
    \textbf{TESS (Ours)} & VGG-9 & \color{green}Yes & 20 & 16 & $98.56\pm0.31\%$ & $22.65$ & $2.26$\\ 
    \midrule
    \multicolumn{8}{c}{CIFAR10 } \\ \midrule
    Hybrid Training\cite{Rathi2020} & VGG-11 & \color{red}No & 250 & 128 & $92.22\%$ & - & -\\ 
    OTTT\textsubscript{A}\cite{Xiao2022OnlineNetworks} & VGG-9 & Partial (in time) & 6 & 128 & $93.52\pm0.06\%$ & - & -\\ 
    \textbf{BPTT (baseline)} & VGG-9 & \color{red}No & 6 & 128 & $92.55\pm0.06\%$ & $3623.90$ & $6.83$\\ 
    \textbf{S-TLLR (baseline)} & VGG-9 & Partial (in time) & 6 & 128 & $91.88\pm0.28\%$ & $3623.90$ & $2.27$\\ 
    \textbf{TESS (ours)} & VGG-9 & \color{green}Yes & 6 & 128 & $92.55\pm0.16\%$ & $5.48$ & $2.27$\\ 
    \midrule
    \multicolumn{8}{c}{CIFAR100} \\ \midrule
    Hybrid Training\cite{Rathi2020} & VGG-11  & \color{red}No & 250 & 128 & $67.87\%$ & - & -\\ 
    OTTT\textsubscript{A}\cite{Xiao2022OnlineNetworks} & VGG-9 & Partial (in time) & 6 & 128 & $71.05\pm0.04\%$ & - & -\\ 
    \textbf{BPTT (baseline)} & VGG-9 & \color{red}No  & 6 & 128 & $69.28\pm0.37\%$ & $3624.18$ & $6.83$\\ 
    \textbf{S-TLLR (baseline)} & VGG-9 & Partial (in time)  & 6 & 128 & $68.00\pm0.71\%$ & $3624.18$ & $2.27$ \\ 
    \textbf{TESS (ours)} & VGG-9 & \color{green}Yes  & 6 & 128 & $70.00\pm0.34\%$ & $17.64$ & $2.27$\\ 
    \bottomrule
    \multicolumn{8}{l}{\small{$^1$: Previous studies' accuracy values are provided as reported in their respective original papers.}}\\
    \multicolumn{8}{l}{\small{$^2$: \# MAC and Memory are estimated for a batch size of 1 with equations presented in Section~\ref{sec:computation_req}.}}\\
    \multicolumn{8}{l}{\small{$^3$: These studies used an input resolution of $128\times128$ for the IBM DVS Gesture dataset.}}
    \end{tabular}
    }
    \end{table*}

    \subsection{Results}
    This subsection presents the results of using TESS across the four datasets, highlighting its performance on image classification tasks and its sensitivity to hyperparameters through ablation studies.
        \subsubsection{Ablation Studies on Eligibility Traces}
        First, we examine the effect of the $\alpha_{\text{post}}$ parameter to include or exclude the non-causal contribution on the learning process.  
        To assess the influence of this parameter, we set $\alpha_{\text{post}}$ to values of $-1$ (negative inclusion), $+1$ (positive inclusion), and $0$ (exclusion).  
        We then trained a VGG-9 model on the four datasets described in Section~\ref{sec:datasets}, performing five independent trials. 
        The mean and standard deviation of the results are reported in Table~\ref{table:ablation_alpha_post}.  
    
        From the results, we observed that, with the exception of the CIFAR10-DVS dataset, the positive inclusion of $\alpha_{\text{post}}$ improves model performance across all other datasets, with gains ranging from $0.23$ to $1.81$ accuracy points compared to when the parameter is excluded.  
        In contrast, the negative inclusion of $\alpha_{\text{post}}$ provides a performance improvement only for the IBM DVS Gesture dataset.  
        
        Beyond performance, note that the inclusion of the $\alpha_{\text{post}}$ parameter also impacts the memory usage of TESS.  
        As discussed in Section~\ref{sec:mem_cost}, when $\alpha_{\text{post}} \neq 0$, additional memory must be allocated for storing the trace of output spikes, $\vh^{(l)}[t]$.  
        Thus, while the inclusion of $\alpha_{\text{post}}$ can enhance model performance, it effectively doubles the memory requirements of the algorithm, allowing for a trade-off between performance and memory usage.

        \subsubsection{Performance on Image Classification}
        In this subsection, we compare the performance of TESS with that of previously reported methods, including non-local, partially local, and fully local learning approaches, on the four datasets described in Section~\ref{sec:datasets}.  
        The results are presented in Table~\ref{table:performance}.
        
        On CIFAR10-DVS, TESS performs on par with prior methods, including those based on backpropagation and temporally local approaches such as \cite{Xiao2022OnlineNetworks, Apolinario2025S-TLLR:Networks}, with a maximum accuracy drop of $2.27\%$, except for \cite{Deng2022TemporalRe-weighting}.  
        A similar trend is observed on CIFAR10 and CIFAR100, where TESS demonstrates a maximum accuracy drop of approximately $1\%$.  
        In contrast, on the DVS Gesture dataset, TESS achieves the best performance among all previously reported methods, including non-local approaches, despite being fully local.  
        Notably, TESS outperforms \cite{Kaiser2020SynapticDECOLLE}, the other fully local method, by approximately $3$ accuracy points.  
        These results highlight the capability of TESS to train models in a fully local manner while achieving performance comparable to or better than non-fully local methods.

        It is worth noting that previous works may differ in experimental implementations, introducing variances that are challenging to quantify in the final results.  
        To address this, we established two baselines for each dataset using BPTT and S-TLLR \cite{Apolinario2025S-TLLR:Networks}, ensuring consistent model implementations, data preprocessing, and hyperparameter settings.  
        Relative to these baselines, TESS outperforms S-TLLR on DVS Gesture, CIFAR10, and CIFAR100 while performing on par with or slightly better than BPTT.  
        The only exception is on CIFAR10-DVS, where TESS lags behind BPTT by $1.4$ and S-TLLR by $0.14$ accuracy points.  
        Furthermore, TESS achieves these results while significantly reducing the computational cost of generating learning signals, with a reduction in MAC operations of $205-661\times$ thanks to its local learning signal generation.  
        Similarly, TESS reduces memory usage by a factor of $3-10\times$ compared to BPTT.  
        
        These findings clearly demonstrate the ability of TESS to train SNN models with accuracy comparable to BPTT while dramatically reducing computational and memory requirements.  
        This makes TESS a highly suitable candidate for enabling learning on low-power devices with constrained resources.   

\section{Conclusions and Perspectives}
We introduced TESS, a temporally and spatially local learning rule for SNNs, designed to meet the demand for low-power, scalable training on edge devices. 
TESS achieves competitive accuracy with BPTT while reducing memory complexity from $O(TLn)$ to $O(Ln)$ and time complexity from $O(TLn^2)$ to $O(LCn)$, making it highly efficient for resource-constrained systems. 
Inspired by biological mechanisms like eligibility traces, STDP, and neural activity synchronization, TESS assigns temporal and spatial credits locally, eliminating the need for global information flow. 
Experiments demonstrate that TESS has accuracy on par with BPTT on datasets such as CIFAR10, CIFAR100, and IBM DVS Gesture, while losing only $1.4$ accuracy points on CIFAR10-DVS.
Moreover, it significantly reduces computational costs, with $205–661\times$ fewer MACs and $3–10\times$ lower memory usage. 
By leveraging its local learning paradigm, TESS offers a scalable, energy-efficient alternative for training SNNs on edge devices, enabling real-time applications with minimal resource demands. 
Its design is particularly promising for low-power on-device learning hardware, where spatiotemporal locality is critical for efficiency \cite{Frenkel2022ReckOn:Timescales}.



\bibliographystyle{IEEEtran}
\bibliography{IEEEabrv, ./references.bib}

\end{document}